\documentclass[11pt]{article}
\usepackage[linesnumbered,ruled,vlined]{algorithm2e}

\usepackage[utf8]{inputenc}

\usepackage[T1]{fontenc}    
\usepackage[table]{xcolor}
\usepackage{url}            
\usepackage{float}
\usepackage{csquotes}
\definecolor{dred}{rgb}{0.6,0,0}
\definecolor{dpurple}{HTML}{A020F0}
\definecolor{dblue}{rgb}{0,0,0.6}
\definecolor{hlcolor}{rgb}{1,1,0.8}

\usepackage{graphicx}
\usepackage{caption}
\usepackage{subcaption}
\usepackage{amsthm}
\usepackage{amsmath,amssymb,bm}
\usepackage{lineno}
\usepackage{qting}

\usepackage{xr-hyper}
\usepackage[hidelinks]{hyperref}

\usepackage[nameinlink]{cleveref}
\usepackage{booktabs}       
\usepackage{amsfonts}       
\usepackage{nicefrac}       
\usepackage{microtype}      
\usepackage{soul}
\usepackage{authblk}
\usepackage{lmodern}
\usepackage{tabularx}
\usepackage{braket}
\usepackage{libertine}

\usepackage[page]{appendix}

\sethlcolor{hlcolor}
\Crefname{equation}{Equation}{Equations}
\Crefname{figure}{Figure}{Figures}
\Crefname{table}{Table}{Tables}
\crefname{section}{Section}{Sections}
\creflabelformat{equation}{#2#1#3}
\crefrangelabelformat{equation}{#3#1#4-#5#2#6}

\renewcommand\b\bm
\usepackage{titlesec}
\titlespacing\section{0pt}{3pt plus 0pt minus 1pt}{1pt plus 0pt minus 1pt}
\titlespacing\subsection{0pt}{2pt plus 0pt minus 1pt}{0pt plus 0pt minus 1pt}
\titlespacing\subsubsection{0pt}{2pt plus 0pt minus 1pt}{0pt plus 0pt minus 1pt}

\titleformat{\section}{\normalfont\large\bfseries}{\thesection}{1em}{}
\titleformat{\subsection}{\normalfont\normalsize\bfseries}{\thesubsection}{1em}{}
\titleformat{\subsubsection}{\normalfont\normalsize\bfseries}{\thesubsubsection}{1em}{}

\definecolor{changecol}{rgb}{0,0.5,0} 
\definecolor{responsecol}{rgb}{0,0,0.8}

\def\q#1#2{{\color[rgb]{0,0.6,0}#2}} 

\usepackage[utf8]{inputenc}
\usepackage[a4paper,bindingoffset=0.2in,left=0.5in,right=0.7in,top=1in,bottom=1in,footskip=.25in]{geometry}
\usepackage{xr-hyper}
\usepackage{url,lineno,microtype,subcaption}
\usepackage[onehalfspacing]{setspace}
\usepackage{mathrsfs}
\usepackage{authblk}
\usepackage[hidelinks]{hyperref}
\usepackage{libertine}

\usepackage{multirow}
\usepackage{graphicx}
\usepackage{subdepth}
\usepackage{setspace}
\usepackage{amsmath,amssymb,mathtools, amsthm}
\DeclareMathOperator{\Tr}{Tr}
\usepackage{bm}
\usepackage{bbm}
\usepackage{comment} 
\usepackage{framed}
\usepackage{tabularx}
\usepackage{booktabs}
\usepackage{pifont}
\newcommand{\cmark}{\ding{51}}%
\newcommand{\xmark}{\ding{55}}%
\newcommand{\smark}{\ding{93}}%
\newcommand{\amark}{\ding{59}}%
\newcommand{\dmark}{\ding{70}}%
\graphicspath{{figures/}}
\usepackage{caption}
\usepackage[numbers,super,sort&compress]{natbib}

\usepackage{titlesec}
\titleformat{\subsubsection}{\bfseries\large}{\thesubsubsection}{1em}{\itshape}
\graphicspath{{figures/}}
\AtBeginDocument{%
}
\makeatletter
\newcommand*{\addFileDependency}[1]{
  \typeout{(#1)}
  \@addtofilelist{#1}
  \IfFileExists{#1}{}{\typeout{No file #1.}}
}
\makeatother

\newcommand*{\myexternaldocument}[1]{%
    \externaldocument{#1}%
    \addFileDependency{#1.tex}%
    \addFileDependency{#1.aux}%
}
\usepackage{xcolor}

\myexternaldocument{supplementary}

\title{Deep Learning without Weight Symmetry}

\author[1]{Li Ji-An}
\author[2, *]{Marcus K. Benna}
\date{}

\affil[1]{\small Neurosciences Graduate Program, University of California San Diego, La Jolla, CA 92093}
\affil[2]{\small Department of Neurobiology, School of Biological Sciences, University of California San Diego, La Jolla, CA 92093}
\affil[*]{\small Corresponding author: mbenna@ucsd.edu}

\begin{document}

\maketitle

\begin{abstract}
Backpropagation, a foundational algorithm for training artificial neural networks, predominates in contemporary deep learning. Although highly successful, it is widely considered biologically implausible, because it relies on precise symmetry between feedforward and feedback weights to accurately propagate gradient signals that assign credit. The so-called \textit{weight transport problem} concerns how biological brains \textit{learn to align} feedforward and feedback paths while avoiding the non-biological transport of feedforward weights into feedback weights.
To address this, several credit assignment algorithms, such as feedback alignment and the Kollen-Pollack rule, have been proposed. While they can achieve the desired weight alignment, these algorithms imply that if a neuron sends a feedforward synapse to another neuron, it should also receive an identical or at least partially correlated feedback synapse from the latter neuron, thereby forming a bidirectional connection. However, this idealized connectivity pattern contradicts experimental observations in the brain, a discrepancy we refer to as the \textit{weight symmetry problem}.
To address this challenge posed by considering biological constraints on connectivity, we introduce the Product Feedback Alignment (PFA) algorithm. We demonstrate that PFA can eliminate explicit weight symmetry entirely while closely approximating backpropagation and achieving comparable performance in deep convolutional networks. Our results offer a novel approach to solve the longstanding problem of credit assignment in the brain, leading to more biologically plausible learning in deep networks compared to previous methods.
\end{abstract}

\section*{Introduction}
Both artificial and biological neural networks must orchestrate complex synaptic weight updates in order to improve task performance. The correct organization of these weight updates, often referred to as the credit assignment problem, becomes even more challenging in deeper neural networks. Over the past decades, the error backpropagation (BP) algorithm has revolutionized contemporary deep learning \cite{rumelhart1986learning}, serving as a fundamental algorithm for training deep artificial neural networks. 

Despite its success, BP is commonly considered biologically implausible. Although recent proposals have partially addressed many biological implausibilities, such as nonlocal plasticity, multiple separate learning phases, and non-biological error representations \cite{whittington2017approximation, sacramento2018dendritic, payeur2021burst,alonso2021tightening}, an important limitation related to the alignment between the forward and backward passes persists. In BP, the forward pass generates output predictions by feeding input data through the network, layer by layer, while the backward pass propagates error signals from the output layer throughout the network in the reverse direction. Following the chain rule of gradient descent, the feedback weights ($W^T$) in the backward pass are precisely symmetric to the feedforward weights ($W$) in the forward pass in order to accurately transmit error signals that match the gradients of the cost function. However, there is no evidence of any biological mechanism capable of directly transporting (i.e., copying) each feedforward weight into the feedback paths --- such a process would require a neuron to track the status of every outgoing synapse.
This highlights the longstanding \textit{weight transport problem} \cite{rumelhart1986learning, stork1989backpropagation, crick1989recent, lillicrap2020backpropagation}: how does the brain perform credit assignment while avoiding the non-biological transport? 

To address this problem, several algorithms have been proposed, in which the feedback weights are either fixed (e.g., feedback alignment), or learned to approximate the transpose of the feedforward weights (e.g., Kollen-Pollack algorithm). However, although these algorithms resolve the issue of biologically implausible weight transport, they still lead to a deeper, less-discussed problem --- (approximate) weight symmetry that contradicts neurobiological measurements. We formulate this \textit{weight symmetry problem} in detail below.

\begin{figure}[htbp]
\centering
\includegraphics[width=\textwidth]{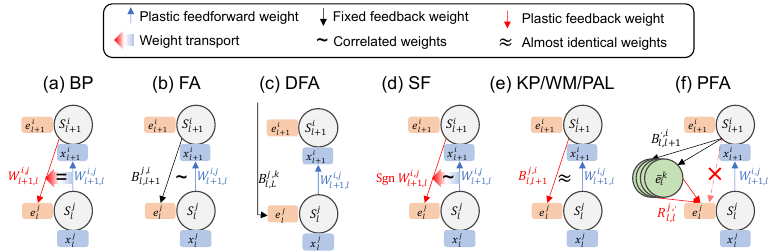}
\caption{
\textbf{Comparison of biologically plausible credit assignment algorithms for multilayer networks.} In the forward pass, the feedforward weight $W^{i,j}_{l+1,l}$ transmits predictions (denoted by $x$) from $S_l^j$ to $S_{l+1}^i$. In the backward pass, all algorithms use a feedback path to propagate errors (denoted by $e$). (a) In BP (backpropagation), the feedback weight is plastic and transported from the feedforward weight $W^{i,j}_{l+1,l}$. (b) In FA (feedback alignment), the feedback weight $B^{j,i}_{l,l+1}$ is kept fixed. (c) In DFA (direct feedback alignment), the fixed feedback weight $B^{j,k}_{l,L}$ projects from $S_L^k$ to $S_l^j$. The weight symmetry problem exists in the final layer. (d) In SF (sign-concordant feedback), the feedback weight is plastic with the sign transported from the feedforward weight ($\text{Sgn}\ W^{i,j}_{l+1,l}$). (e) In KP (Kollen-Pollack algorithm), WM (weight mirror), and PAL (phaseless alignment learning), the feedback weight is plastic and updated by a local plasticity rule, avoiding the transport. (f) In PFA (product feedback alignment), the error signal is propagated through an additional population $\bar e^k_l$. Unlike other algorithms in (a-e), PFA resolves the weight symmetry problem by avoiding the direct feedback synapse from $S_{l+1}^i$ to $S_l^j$.} 
\label{fig:algorithm}
\end{figure}

In BP (Fig.~\ref{fig:algorithm}a), the feedforward synaptic weight $W_{l+1,l}^{i,j}$ from neuron $S_l^j$ (the $j$-th neuron in layer $l$) to neuron $S_{l+1}^i$ is updated by (an additive plasticity term proportional to) the negative gradient of the loss function $\mathcal{L}$ with respect to $W_{l+1,l}^{i,j}$. This term is computed as the product of the forward prediction $x_l^j$ and the backward error signal $e_{l+1}^i$ (the negative gradient of  $\mathcal{L}$ with respect to $x_{l+1}^i$). For BP to be biologically plausible in the brain \cite{whittington2017approximation, sacramento2018dendritic}, the error signals must be locally available for the feedforward weights, meaning that $e_{l+1}^i$ should be transmitted to the neuron $S_{l+1}^i$ that represents $x_{l+1}^i$ (e.g., $e_{l+1}^i$ and $x_{l+1}^i$ represented by different dendritic compartments). 
Because chemical synapses in the brain transmit information (carried by presynaptic spikes) only in one direction (from axon to dendrite), the BP algorithm implies that, if there is a feedforward synapse $W_{l+1,l}^{i,j}$ from $S_l^j$ to $S_{l+1}^i$, there must also exist a feedback synapse with identical strength from $S_{l+1}^i$ to $S_l^j$. However, this symmetric weight pattern is not observed in the biological brain, contradicting the empirical synaptic connectivity and efficacy data \cite{song2005highly}. For example, in rat visual cortex, two connected neurons $S_1$ and $S_2$ are either unidirectionally connected, with a probability of 69\% ($S_1\rightarrow S_2$ or $S_2\rightarrow S_1$), or bidirectional connected, with a probability of 31\% ($S_1\rightarrow S_2$ and $S_2\rightarrow S_1$) \cite{song2005highly}. Even for bidirectionally connected neurons,  $W_{S_1\rightarrow S_2}$ and $W_{S_2\rightarrow S_1}$ (representing synaptic strength) are only modestly correlated ($r\approx 0.36$) \cite{song2005highly} (see Appendix B). These observations are in striking contrast to the bidirectional connection and perfect correlation ($r=1$) in the symmetric weight connectivity assumed by BP. This discrepancy thus calls for alternative explanations involving credit alignment algorithms that align with biological mechanisms. 

In this study, we first review existing algorithms proposed to resolve the weight transport problem and demonstrate their failure to address the weight symmetry problem. We then propose the Product Feedback Alignment (PFA) algorithm, which completely avoids \emph{explicit} weight symmetry by relying on \emph{indirect} alignment between forward and backward paths, using an additional population of neurons in the backward paths. Specifically, the feedforward weights $W$ align with the product of a pair of feedback weight matrices $R$ and $B$ (such that $W\propto (RB)^T$), thus any pair of connected neurons is only unidirectionally connected. We then characterize the BP-approximating properties of PFA and empirically show that PFA can achieve  BP-level performance in deeper networks, convolutional layers, and more challenging datasets such as CIFAR10 and ImageNet. 
Our results offer a novel formulation and solution to the longstanding problem of biological credit assignment, supporting the possibility that the biological brain naturally implements BP-like algorithms.

\begin{table}[]
\begin{tabular}{ccccccccc}
\hline
                                               & BP & FA & DFA & SF & KP & WM & PAL & \textbf{PFA} \\
\hline
No need to transport weight sign           & \xmark  & \cmark  & \cmark   & \xmark  & \cmark  & \cmark  & \cmark  & \textbf{\cmark}   \\
No need to transport weight magnitude      & \xmark  & \cmark  & \cmark   & \cmark  & \cmark  & \cmark  & \cmark  & \textbf{\cmark}   \\
No separate feedback weight learning phase & \cmark  & \cmark  & \cmark   & \amark  & \cmark  & \xmark  & \cmark  & \textbf{\cmark}   \\
No explicit weight symmetry after training     & \xmark  & \smark  & \smark   & \smark  & \xmark  & \xmark  & \smark  & \textbf{\cmark}   \\
Accurate approximation to BP (path alignment)                 & \cmark  & \xmark  & \xmark   & \xmark  & \cmark  & \cmark  & \xmark  & \textbf{\cmark}   \\
BP-level task performance                      & \cmark  & \xmark  & \xmark   & \dmark  & \cmark  & \cmark  & \dmark  & \textbf{\cmark}   \\
\hline
\end{tabular}
\caption{Detailed comparison of algorithms. BP: backpropagation. FA: feedback alignment. DFA: direct feedback alignment. SF: sign-concordant feedback. KP: Kollen-Pollack algorithm. WM: weight mirror. PAL: phaseless alignment learning. PFA: product feedback alignment. \amark: It is unclear how the feedback weights in SF can be learned in a biologically plausible way. \smark: these algorithms reduce, but do not fully eliminate explicit weight symmetry. \dmark: These algorithms significantly outperform FA and DFA, but still underperform compared to BP in more challenging tasks (CIFAR10 for PAL and ImageNet for SF).}
\label{tab:comparison}
\end{table}
\section*{Results}

\subsection*{Existing credit assignment algorithms fail to address the weight symmetry problem}

Consider a fully-connected multilayer neural network with depth $L$, mapping the input $\bm x_0$ to the output $\bm x_L$. 
In the forward pass, the activation of layer $l+1$ (with $N_{l+1}$ neurons), denoted as $\bm x_{l+1}$, is determined by
\begin{equation}
    \bm x_{l+1}=\sigma_{l+1}(\bm a_{l+1})= \sigma_{l+1}(W_{l+1,l} \bm x_{l} + \bm b_{l+1})\ ,
\end{equation}
where $\bm a_{l+1}$ is the preactivation, $\sigma$ is the activation function ($\sigma_l=\sigma$ for $l<L$ and $\sigma_L=\mathrm{Id}$), $W_{l+1,l}$ is the $N_{l+1}\times N_l$ feedforward weight, and $\bm b_{l+1}$ is the bias. $W_{l+1,l}^{i,j}$ and $b_{l+1}^i$ are scalars representing individual synaptic weight and bias, respectively.
For a training dataset consisting of data points $(\bm x_0, \bm y)$, the set of parameters ${W_{l+1,l}}$ and $\bm b_l$ are trained to minimize the loss function $\mathcal{L}$ that measures the difference between layer $L$'s output $\bm x_L$ and the target output $\bm y$ (e.g., the cross-entropy). The loss at the output layer directly provides the teaching signal (error) $\bm e_L$ that is locally available at $\bm x_L$, defined as the negative gradient $\bm e_L=-\partial \mathcal{L} / \partial \bm a_L$. 

\paragraph{Backpropagation (BP).} In the backward pass  (Fig.~\ref{fig:algorithm}a), the error (i.e., gradient) at layer $\bm x_l$ is iteratively propagated as 
\begin{equation}
    \bm e_l^{\rm BP}=\sigma'(\bm a_l) \odot W_{l+1,l}^T  \bm e_{l+1}^{\rm BP}\ , 
\end{equation}
where $\odot$ is the Hadamard product. Subsequently, the feedforward weight $W_{l+1,l}$ is updated by 
\begin{equation}
    \Delta W_{l+1,l}=\eta  \bm e_{l+1}^{\rm BP} \bm x_l^T\ , 
\end{equation}
where $\eta$ is the learning rate. Here $e_{l+1}^i$ ($i$-th component of $\bm e_{l+1}$) is locally available at the neuron $S_{l+1}^i$ with activation $x_{l+1}^i$, and the update of $W_{l+1,l}^{i,j}$ (forward synaptic weight between $S_{l+1}^{i}$ and $S_{l}^{j}$) relies only on locally available information ($e_{l+1}^{i}$ available in $S_{l+1}^{i}$ and $x_{l}^{j}$ available in $S_{l}^{j}$), which is biologically plausible. The aspect that lacks biological plausibility is the backpropagation  of errors via $W_{l+1,l}^T$, since the same synaptic weight $W_{l+1,l}^{i,j}$ is used twice in both the forward and backward passes. This requires the non-biological transport of feedforward weights into feedback paths.

\paragraph{Feedback alignment (FA).} In an effort to resolve the weight transport problem, FA uses random fixed feedback weights $B_{l,l+1}^{\rm FA}$ ($N_l\times N_{l+1}$) \cite{lillicrap2016random}, which are independent of the feedforward weights. it was demonstrated that FA can transmit useful error signals to upstream layers. Specifically, the error at the layer $\bm x_l$ is computed as $\bm e_l^{\rm FA}=\sigma'(\bm a_l) \odot B^{\rm FA}_{l,l+1} \bm e_{l+1}^{\rm FA}$. However, FA struggles to match BP's performance in more advanced network architectures and in more challenging tasks, including deeper networks, convolutional layers, and large-scale image datasets (e.g., CIFAR10, ImageNet) \cite{bartunov2018assessing}. Another FA variant was proposed by meta-learning the plasticity rule of feedforward weights to improve FA for online learning and low-data regimes \cite{shervani2023meta}, but it still significantly underperforms relative to BP. Further, FA leads to a learning process \cite{lillicrap2016random} whereby the feedforward weights $W_{l+1,l}$ gradually become correlated with the feedback weights $B_{l,l+1}^T$, i.e., $W_{l+1,l}\sim B_{l,l+1}^T$, leading to an approximate weight symmetry to an extent that contradicts biological observations. 

\paragraph{Direct feedback alignment (DFA).} DFA calculates the error at the layer $\bm x_l$ using $\bm e_l^{\rm DFA}=\sigma'(\bm a_l) \odot B_{l,L}^{\rm DFA} \bm e_{L}$ via a fixed random feedback weight $B_{l,L}^{\rm DFA}$ ($N_l\times N_L$) that projects from the final layer. DFA only has correlated weights in the final layer ($W_{L,L-1}\sim B_{L-1,L}^T$), alleviating the weight symmetry problem. However, DFA similarly failed to reach good performance for deeper networks, convolutional layers, and more challenging tasks  \cite{bartunov2018assessing}. 

\paragraph{Sign-concordant feedback (SF).} Since fixed feedback weights hinder task performance, other proposals have explored various methods for updating feedback weights. In SF, the sign of feedforward weights are transported into backward paths both at initialization and during training \cite{liao2016important}, i.e., $B^{\rm SF}_{l,l+1}=\text{Sign}(W^T_{l+1,l})$, and the error $\bm e_l$ is computed as $\bm e_l^{\rm SF}=\sigma'(\bm a_l) \odot B^{\rm SF}_{l,l+1} \bm e_{l+1}^{\rm SF}$. SF significantly outperforms FA and DFA, though still shows a significant performance gap in more complex tasks like ImageNet \cite{xiao2018biologically, moskovitz2018feedback,sanfiz2021benchmarking}. Overall, it remains unclear whether a biologically plausible mechanism can effectively transport the sign of the synaptic weight from the forward path to the corresponding synaptic weight in the backward path. Furthermore, the feedforward and feedback weights are strongly correlated \cite{sanfiz2021benchmarking}, and thus SF does not address the weight symmetry problem.

\paragraph{Kollen-Pollack (KP), weight mirror (WM), phaseless alignment learning (PAL).} KP \cite{akrout2019deep} uses a feedback weight update symmetric to the feedforward weight update ($\Delta B^{\text{KP}}_{l,l+1}=\eta  \bm x_l (\bm e_{l+1}^{\text{KP}})^T$) and includes weight decay.  
It is shown to closely approximate BP, achieving similar performance. However, KP leads to explicit weight symmetry. In WM \cite{akrout2019deep}, the feedback weights are updated to track the feedforward weights, by injecting random noise into neurons during multiple learning phases (one phase for each layer). WM can reach a performance similar to BP, although the biological  feasibility of layer-specific learning phases and ``bias blocking'' (setting bias to zero) during the mirror mode remains to be established. However, even starting from asymmetric initializations,  WM ultimately leads to symmetric feedforward and feedback weights. PAL  \cite{max2023learning} eliminates the need for the additional mirror mode in WM, but again deviates from BP's dynamics, showing a significant performance gap compared to BP. The learned feedback weights in PAL are still strongly correlated with the feedforward weights.

\paragraph{Summary.} KP and WM can achieve a BP-level task performance but converge to a configuration with almost exact weight symmetry. FA, DFA, SF, and PAL alleviate the weight symmetry issue (although their weight configurations are still more aligned than and arguably incompatible with biological observations), but significantly sacrifice task performance. None of these algorithms manages to achieve a BP-level performance while completely avoiding explicit weight symmetry.

\subsection*{Product feedback alignment resolves the weight symmetry problem}
Motivated by the diversity of neuronal populations observed in the brain, we propose augmenting the backward pass with additional populations to address this weight symmetry problem. We formalize the question as follows:
Given an intermediate population $\bm {\bar e}_l$ (with $\bar N_{ l}$ neurons) for each layer $l$, along with a set of weights $B_{l,l+1}^{\rm PFA}$ ($\bar N_{l}\times N_{l+1}$) projecting from $\bm e_{l+1}$ to $\bm {\bar e}_l$ and weights $R_{l,l}$  ($N_l \times \bar N_l$) projecting from $\bm{\bar e}_{l}$ to $\bm e_l$, what biologically plausible plasticity rules of 
$B_{l,l+1}^{\rm PFA}$ and $R_{l,l}$ can approximate BP?

This formulation leads to the PFA solution (Fig.~\ref{fig:algorithm}f). Similarly to BP, the feedforward weight matrix $W_{l+1,l}$ is updated by $\Delta W_{l+1,l}=\eta \bm e^{\rm PFA}_{l+1}\bm x_l^T$.
In the backward pass, $\bm{\bar e}_l$ is computed as
\begin{equation}
    \bm{\bar e}_l=B_{l,l+1}^{\rm PFA} \bm e_{l+1}^{\rm PFA}\ , 
\end{equation}
using a fixed random weight matrix $B_{l,l+1}^{\rm PFA}$. Subsequently, the error $\bm e_l^{\rm PFA}$ is calculated as 
\begin{equation}
    \bm e_l^{\rm PFA}=\sigma'(\bm a_l) \odot R_{l,l} \bm{\bar e_{l}}\ , 
\end{equation}
where $R_{l,l}$ is plastic, updated as
\begin{equation}
    \Delta R_{l,l}=\eta \bm x_l \bm{\bar e}_{l}^T\ . 
\end{equation}

Note that the update of the synaptic weight $R_{l,l}^{j,k}$ only relies on locally available information $x_{l}^{j}$ and $\bar e_{l}^{k}$. The backward path between layers $l+1$ and $l$ consists of a pair of feedback weights $B^{\rm PFA}_{l,l+1}$ and $R_{l,l}$. Weight decay is applied for both $W_{l+1,l}$ and $R_{l,l}$, such that the influence of initializations gradually diminishes. We will elucidate the \emph{implicit} alignment mechanism of PFA below. 

With this additional population $\bm {\bar e}_l$, PFA is more consistent with the neuronal recordings by eliminating the feedback synapse from neuron $S^i_{l+1}$ to $S^j_{l}$. We note that PFA generalizes FA and KP: when the weight decay and learning rate for $R_{l,l}$ are set to zero, PFA reduces to FA; when $B^{\rm PFA}_{l,l+1}$ is set to an identity matrix (with $\bar N_{l}=N_{l+1}$), PFA reduces to the KP algorithm.

\subsection*{PFA approximates BP in MNIST handwritten digit classification}
\label{sec:mnist}

\begin{figure}[htbp]
\centering
\includegraphics[width=\textwidth]{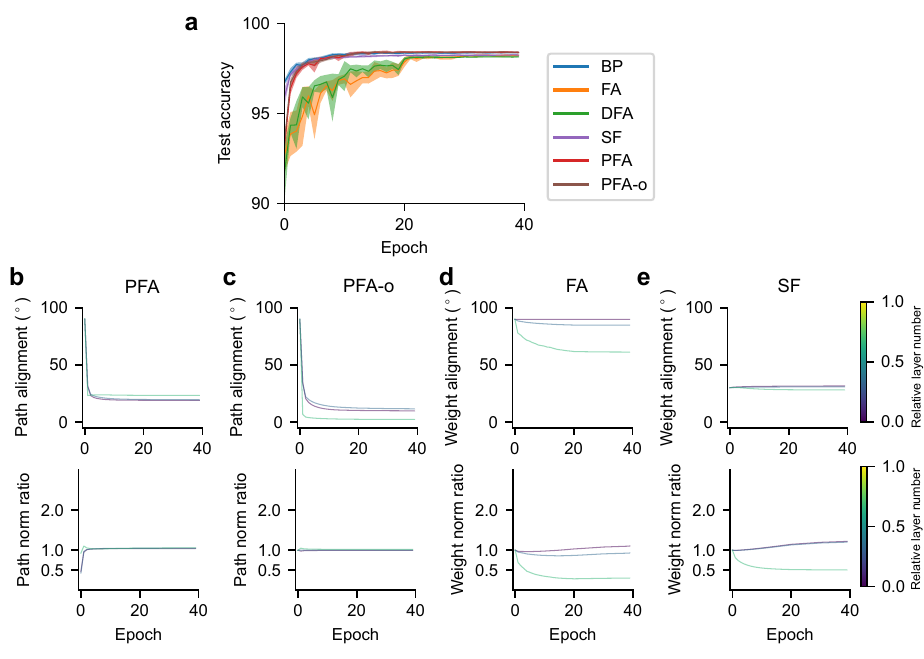}
\caption{
\textbf{Characterization of learning algorithms for two-hidden-layer feedforward networks trained to classify  MNIST digit images.} (a) Task performance. Shaded regions show standard deviations across 5 seeds.  PFA and PFA-o curves are almost overlapping with the BP curve, suggesting a close approximation. (b-e) Backward-forward weight alignment for FA/SF, and path alignment for PFA/PFA-o (top). Backward-forward weight norm ratio for FA/SF, and path norm ratio for PFA/PFA-o (bottom). } 
\label{fig:mnist}
\end{figure}

We compare these learning algorithms in training a two-hidden-layer feedforward network (with layer sizes 784-512-512-10) with ReLU activation on the MNIST handwritten digit dataset (using the BioTorch framework \cite{sanfiz2021benchmarking}). The expansion ratio ($1/\lambda=\bar N_{l}/N_{l+1}$) in PFA is set to 10. See Appendix \ref{sec:training_details} for training details. We found that PFA (with randomly sampled elements in $B_{l,l+1}$), PFA-o (with a semi-orthogonal matrix $B_{l,l+1}$), and SF reach a test performance similar to BP, slightly outperforming FA and DFA (Fig.~\ref{fig:mnist}a).

To track the learning process on MNIST, we recorded two types of key metrics throughout training. The first type of metric, backward-forward \emph{weight alignment} \cite{sanfiz2021benchmarking} angle $\angle(\textbf{Vec}(W_{l+1,l}^T), \textbf{Vec}(B_{l,l+1}))$, quantifies the angle between the feedforward weights and the corresponding feedback weights, where $(\cdot,\cdot)$ denotes the normalized inner products. A larger weight alignment angle (closer to $90^\circ$) is more consistent with brain recordings, since it indicates less weight symmetry. In PFA, the backward-forward weight alignment angle vanishes into $90^\circ$ (i.e., there is no direct feedback weight, hence no chance of weight symmetry). We thus define the backward-forward \emph{path alignment} angle $\angle(\textbf{Vec}(W_{l+1,l}^T), \textbf{Vec}(R_{l,l} B_{l,l+1}))$ for PFA that generalizes the weight alignment angle. Crucially, a small path alignment angle for PFA is related to successful credit assignment (approximating BP) but does not indicate weight symmetry. Our PFA achieved an angle around $18^\circ$ (Fig.~\ref{fig:mnist}b) for all layers in the network after the initial epochs (because we set a large initial weight decay for $R$ and $W$). Due to the orthogonal initialization for $B$ (Fig.~\ref{fig:mnist}c), PFA-o gradually achieved an angle close to $0^\circ$. These low path alignment angles indicate a superior alignment in PFA (but crucially without weight symmetry). 

The second type of metric, backward-forward \emph{weight norm ratio}, defined as $ ||B_{l,l+1}||_2 / ||W_{l+1,l}^T||_2$, is designed to capture the risk of experiencing gradient exploding/vanishing problems observed in FA \cite{moskovitz2018feedback, sanfiz2021benchmarking}. Similarly, the backward-forward weight norm ratio vanishes to zero in PFA. We thus defined the backward-forward \emph{path norm ratio} $ ||R_{l,l} B_{l,l+1}||_2 / ||W_{l+1,l}^T||_2$ for PFA. Our PFA and PFA-o attained a weight ratio near $1$ for all layers in the network -- the ideal ratio akin to BP -- after the initial epochs, suggesting better stability in error propagation over FA (between 0.3 and 1.1) and SF (between 0.5 and 1.2) (Fig.~\ref{fig:mnist}b-e). 

Collectively, these findings show that PFA achieves a close approximation to BP in both task performance and training dynamics.

\subsection*{Alignment mechanism between forward and backward paths in PFA}
\label{sec:alignment}

\begin{figure}[htbp]
\centering
\includegraphics{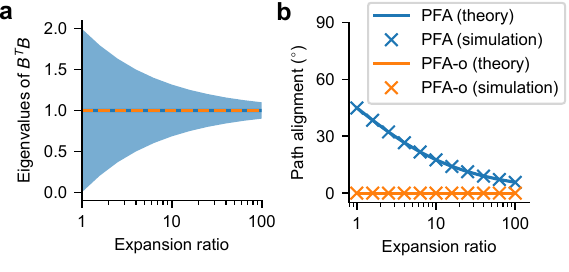}
\caption{\textbf{PFA gradually approximates BP as the expansion ratio increases.}
(a) Eigenvalues of $B^TB$ as a function of the expansion ratio ($1/\lambda$). The shaded region shows the standard deviation of the eigenvalues. (b) Backward-forward path alignment between $W$ and $(RB)^T$ as a function of the expansion ratio ($1/\lambda$). For simulations, we randomly sampled $W$ and set $R^T=BW$ (expected to hold after the effect of the weight initialization has fully decayed). The theoretical predictions from \textbf{Proposition 3} match our simulations and are consistent with the observed path alignment after training. The (invisible) shaded regions show the standard deviations of simulations.} 
\label{fig:BTB_ev}
\end{figure}

We now elucidate the alignment mechanism between the forward and backward paths. During training, the feedforward weights $W_{l+1,l}$ align with the feedback weight matrix product $(R_{l,l} B_{l,l+1})^T$ when the following two requirements are satisfied.

(1) $B_{l,l+1}^T B_{l,l+1}$ approximates an identity matrix. Then the weight updates for $W_{l+1,l}$ and $(R_{l,l}B_{l,l+1})^T$ are aligned:
\begin{equation} 
\label{eq:alignment}
\begin{split}
\Delta (R_{l,l}B_{l,l+1})^T &=B_{l,l+1}^T \Delta R_{l,l}^T=\eta B_{l,l+1}^T \bm{\bar e}_{l}\bm x_l^T=\eta (B_{l,l+1}^T B_{l,l+1}) \bm e_{l+1}\bm x_l^T \\
&\approx \eta \bm e_{l+1}\bm x_l^T=\Delta W_{l+1,l} \ . 
\end{split}
\end{equation}

(2) The influence of the initial values of $W_{l+1,l}$ and $R_{l,l}$, which causes misalignment between $W_{l+1,l}$ and $(R_{l,l}B_{l,l+1})^T$, further diminishes over successive learning epochs (due to weight decay).

Different initializations can ensure that $B_{l,l+1}^T B_{l,l+1}$ approximates an identity matrix. The first approach (``PFA-o'') is to initialize $B_{l,l+1}$ as a semi-orthogonal ($\bar N_{l}\times N_{l+1}$) matrix (with $\bar N_{l} \geq N_{l+1}$).

Alternatively, elements in $B_{l,l+1}$ can be sampled independently from a distribution with a mean of $0$ and a variance of $1/\bar N_{l}$ (``PFA''). This initialization for PFA is easier to implement in the brain than PFA-o, as it is not obvious what biological plasticity rule can learn a precisely semi-orthogonal matrix. Following the Marchenko–Pastur law \cite{bai2010spectral} and assuming the limit $N_{l+1}\rightarrow\infty$ and $\bar N_{l}\rightarrow\infty$ with the ratio $N_{l+1}/\bar N_{l}=\lambda <1$, the eigenvalue density $\mu(v)$ of $B_{l,l+1}^T B_{l,l+1}$ satisfies
\begin{equation} 
\label{eq:MP}
\mu(v)=\frac{1}{2\pi}\frac{\sqrt{(\lambda_+-v)(v-\lambda_-)}}{\lambda v}\mathbbm{1}_{v\in [\lambda_-, \lambda_+]}
\end{equation}
with $\lambda_\pm =(1\pm\sqrt{\lambda})^2$. In the limit $\lambda\rightarrow 0$, $B_{l,l+1}^T B_{l,l+1}$ converges to the identity matrix with probability 1. See Fig.~\ref{fig:BTB_ev}a for a numerical verification.

Formally, we provide three propositions to characterize the BP-approximating properties, with proof in Appendix C:

\paragraph{Proposition 1.} In PFA-o ($\lambda\le 1$) and PFA (in the limit $\lambda\rightarrow 0$), the backward-forward path alignment angle $\angle(\textbf{Vec}(W^T), \textbf{Vec}(RB))$ converges to 0 after learning.

\paragraph{Proposition 2.} In PFA-o ($\lambda\le 1$) and PFA (in the limit $\lambda\rightarrow 0$), the error $\bm e_l^{\rm PFA}$ at layer $l$ is fully aligned to $\bm e_l^{\rm BP}$ ($\bm e_l^{\rm PFA}=\bm e_l^{\rm BP}$) after the path alignment angle converges to 0, and thus the loss function is monotonically decreasing for infinitesimal learning rates.

\paragraph{Proposition 3.} In PFA (with $\lambda>0$), the backward-forward path alignment angle $\angle(\textbf{Vec} (W^T), \textbf{Vec}(RB))$ converges to $\arccos(1/\sqrt{1+\lambda})$ almost surely after learning.

In short, although $W_{l+1,l}$ does not align with $R_{l,l}$ or $B_{l,l+1}$ individually, we still have $W_{l+1,l}^T\approx R_{l,l}B_{l,l+1}$ (see Fig.~\ref{fig:BTB_ev}b for numerical verification) and thus $W_{l+1,l}^T \bm e_{l+1}\approx R_{l,l}B_{l,l+1} \bm e_{l+1}$, signifying a good approximation to BP. Like FA and DFA, PFA obviates the need to transport the feedforward weight (or its sign, as in SF) at initialization or during training, but significantly outperforms FA and DFA. Like KP, PFA employs aligned weight updates for both forward and backward paths. However, unlike KP and WM, PFA does not lead to \emph{explicit} weight symmetry, i.e., forward and backward synapses between any pair of neurons do not share identical weights (in fact we avoid bidirectionally connected pairs of neurons altogether). For a detailed comparison of algorithms, see Table~\ref{tab:comparison}.

\subsection*{PFA closely approximates BP in deep convolutional networks}
While FA and DFA face substantial challenges in deep convolutional networks \cite{moskovitz2018feedback}, we show that our PFA performs well in convolutional layers. Consider a forward convolutional kernel $W_{l+1,l} \in \mathbb{R}^{h\times w \times N_{l+1} \times N_l}$, where $h$ is the kernel height, $w$ is the kernel width, and $N_l$ is the number of channels in the $l$-th layer. The feature-map activation $x_{l+1}^i$ for the $i$-th channel at layer $l+1$ is calculated as
\begin{equation} 
x_{l+1}^i= \sigma (a_{l+1}^i)= \sigma (\sum_{j=1}^{N_l} W_{l+1,l}^{i,j} * x_{l}^j + b_{l+1}^i) \ , 
\end{equation} 
where $W_{l+1,l}^{i,j}$ is the $h\times w$ kernel and $*$ denotes convolution. The BP error (i.e., gradient) in the backward pass for $j$-th channel at the $l$-th  layer is iteratively calculated as 
\begin{equation} 
e_l^{{\rm BP},j}=\sigma'(a_l) \odot \sum_{i=1}^{N_{l+1}}\tilde W_{l+1,l}^{i,j} * e_{l+1}^{{\rm BP},i} \ , 
\end{equation} 
where $\odot$ is the Hadamard product, and $\tilde W$ is the a rotation of $W$ by $180^\circ$ (flipped kernel) \cite{goodfellow2016deep}. The FA error for $j$-th channel at layer $l$ is calculated as $e_l^{{\rm FA},j}=\sigma'(a_l) \odot \sum_{i=1}^{N_{l+1}} B_{l,l+1}^{{\rm FA},j,i} * e_{l+1}^{{\rm FA},i}$, where $B_{l,l+1}^{\rm FA} \in \mathbb{R}^{h\times w \times N_l \times N_{l+1}}$ is the backward kernel. For PFA, the intermediate error for the $k$-th intermediate channel at the $l$-th layer is  
\begin{equation} 
\bar e_l^{k}=\sum_{i=1}^{N_{l+1}} B_{l,l+1}^{k,i} * e_{l+1}^{{\rm PFA},i} \ ,
\end{equation} 
where $B_{l,l+1} \in \mathbb{R}^{1\times 1\times \bar N_{l} \times N_l }$. The PFA error for the $j$-th channel at layer $l$ is calculated as 
\begin{equation} 
e_l^{{\rm PFA},j}=\sigma'(a_l) \odot \sum_{k=1}^{\bar N_{ l}} \tilde R_{l,l}^{j,k} * \bar e_{l}^{k} \ , 
\end{equation} 
where $R_{l,l} \in \mathbb{R}^{h\times w \times N_{l} \times \bar N_{ l}}$. 
The pair of tensors $B_{l,l+1}$ and  $R_{l,l}$ accommodate different convolutional hyperparameters for $W_{l+1,l}$ (e.g., stride, padding, dilation, groups). The weight updates are calculated similarly as $\Delta W_{l+1,l}^{i,j}=\eta e_{l+1}^{i} * x_l^j$ (in BP and PFA) and $\Delta R_{l,l}^{j,k}=\eta x_l^j*\bar e_{l}^{k}$ (in PFA).


\label{sec:conv}
\begin{figure}[htbp]
\centering
\includegraphics[width=1\textwidth]{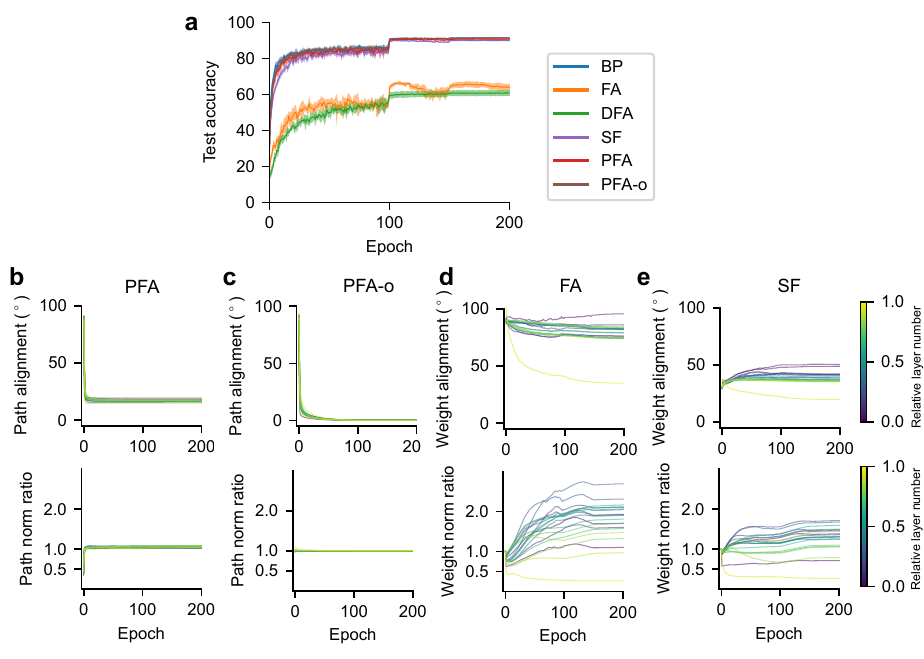}
\caption{
\textbf{Characterization of learning algorithms for ResNet-20 on CIFAR-10.} (a) Task performance. Shaded regions show standard deviations across 5 seeds. PFA and PFA-o curves are almost overlapping with the BP curve, suggesting a close approximation. (b-e) Backward-forward weight alignment for FA/SF, and path alignment for PFA/PFA-o (top). Backward-forward weight norm ratio for FA/SF, and path norm ratio for PFA/PFA-o (bottom).
}
\label{fig:cifar10}
\end{figure}

\begin{figure}[htbp]
\centering
\includegraphics{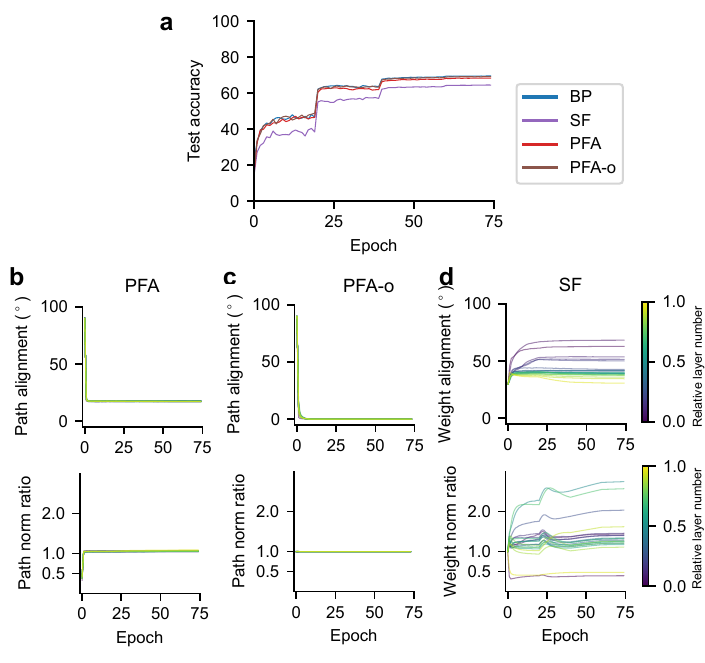}
\caption{
\textbf{Characterization of learning algorithms for ResNet-18 on ImageNet.} (a) Task performance. PFA and PFA-o curves are almost overlapping with the BP curve, suggesting a close approximation. (b-d) Backward-forward weight alignment for SF, and path alignment for PFA/PFA-o (top). Backward-forward weight norm ratio for SF, and path norm ratio for PFA/PFA-o (bottom).
}
\label{fig:imagenet}
\end{figure}
To examine our PFA on deep convolutional networks, we trained ResNet-20 on the CIFAR-10 dataset for these learning algorithms \cite{he2016deep, sanfiz2021benchmarking}. See Appendix \ref{sec:training_details} for training details. The networks trained with BP (Fig.~\ref{fig:cifar10}a) reached an accuracy of 91.64\%, comparable to the performance in the literature (91.37\% in \cite{sanfiz2021benchmarking}). The networks trained with PFA and PFA-o attain a test accuracy comparable to that of BP and SF, significantly surpassing that of FA and DFA (Fig.~\ref{fig:cifar10}a). PFA and PFA-o again achieve a close approximation to BP (Fig.~\ref{fig:cifar10}b-e), indicated by the small angle of path alignment and the close-to-one backward-forward path norm ratio.

We additionally trained ResNet-18 on ImageNet for these algorithms (except FA and DFA, which were previously shown to perform poorly \cite{akrout2019deep, sanfiz2021benchmarking}). See Appendix~\ref{sec:training_details} for training details. The networks trained with BP (Fig.~\ref{fig:imagenet}a) reached an accuracy of 69.69\%, comparable to the performance in the literature (69.61\% in \cite{sanfiz2021benchmarking} and 69.9\% in \cite{akrout2019deep}). Our results (Fig.~\ref{fig:imagenet}a) again show that networks trained with PFA (68.46\%) and PFA-o (69.30\%) achieve a test accuracy comparable to that of BP (69.69\%), in a situation where even SF (64.63\%) significantly underperforms. PFA and PFA-o again achieve a close approximation to BP (Fig.~\ref{fig:imagenet}b-d), as shown by their path alignment and path norm ratio.



\section*{Discussion}
In conclusion, our PFA algorithm approximates BP by fostering implicit alignment between forward and backward paths, rather than relying on explicit weight symmetry. Compared with existing algorithms, PFA effectively resolve the weight symmetry problem. PFA provides a potential explanation for why we do not observe symmetric weights in the brain even though the brain presumably implements credit assignment across multilayer network architectures, serving as a demonstration of feasibility. We showed that PFA achieves a performance comparable to BP in both fully-connected feedforward networks and deeper convolutional networks, even on rather challenging datasets/tasks where other biologically plausible learning algorithms failed. 

Our PFA algorithm utilizes an additional neuronal population $\bm{\bar e}_l$ to transmit errors $\bm{e}_l$ to the population with activation $\bm x_l$. This is broadly consistent with  the brain's diverse neuronal population structure, with predominantly local connections. We have demonstrated that introducing this population allows us to fully remove all bidirectional connections between neurons, leading to a weight alignment angle of $90^\circ$ (i.e., zero weight correlation). To accurately match the weight correlation patterns in the brain, the observed values of weight alignment/correlation (an alignment angle of $78^\circ$ and a correlation of $R\approx 0.2$, estimated in Appendix~\ref{sec:brain_corr}) can be directly achieved by combining different groups of synapses separately learned by PFA (no correlation), FA (partial correlation), and KP (full correlation).
Further, additional populations like the one introduced here  provide extra flexibility and offer the possibility to deal with other biological constraints (such as sparse connections), a direction yet to be fully explored. 

In our simulations, we fixed the feedback weights $B$, and since in PFA (but not PFA-o) each synaptic weight in $B$ was independently sampled, a fairly large expansion ratio was necessary to ensure a close approximation to BP. Exploring biologically plausible plasticity rules for $B$ that could reduce the expansion ratio requirement, potentially leading to a $B^TB$ closer to the identity matrix, is one of our future research directions. We only considered an expansion ratio larger than one ($\lambda<1$) in this study because an expansion ratio smaller than one will reduce the rank of the transmitted errors, similar to the scenario of low-rank gradient approximation \cite{vogels2019powersgd}. 
Further investigation is required to understand the effects of low-rank weight updates. 

The plasticity rules for updating the feedforward weights $W$, which are based on presynaptic activations and postsynaptic errors, are only biologically plausible when the error signal $e_l^i$ is locally available at the corresponding neuron with activation $x_l^i$. Several proposals have been put forward regarding how a single neuron can represent and transmit both forward activations and backward errors without interference. 
The dendritic cortical microcircuit framework posits  that the errors are represented in the apical dendrites while the forward activations are represented in the basal dendrites  \cite{sacramento2018dendritic}. An additional self-prediction path is introduced to facilitate error transmission. Another study, based on burst-dependent plasticity, proposes that the forward activations are represented by the event firing rates, while errors are encoded by burst probabilities \cite{payeur2021burst}, enabling the multiplexing of signals within the same neuron. Our PFA algorithm can be directly integrated  with these proposals, resolving the explicit weight symmetry problem in these frameworks.


\section*{Acknowledgements}
M.K.B was supported by NIH R01NS125298. M.K.B and L.J.-A. were supported by the Kavli Institute for Brain and Mind.



\bibliographystyle{unsrt}
\bibliography{test}

\newpage
\section{Appendix}

\subsection{Training details}
\label{sec:training_details}
The expansion ratio ($1/\lambda=\bar N_{l}/N_{l+1}$) in PFA is set to 10.
All weights in the networks are initialized with the Glorot initialization with unit gain factor \cite{glorot2010understanding}, except for the $B$ weights in PFA, which are determined using the Glorot initialization with gain factor $\sqrt{(1+\lambda)/2}$, ensuring that $B^TB$ approximates an identity matrix (or $B$ is directly set equal to a semi-orthogonal matrix for the ``PFA-o'' variant). 
For optimization, we employed stochastic gradient descent with a momentum of 0.9.

\subsubsection{MNIST}

 The learning rate is $10^{-2}$ for BP, SF, PFA, FA and DFA, multiplied by a factor of 0.5 at the 20th and 30th epoch. The weight decay coefficient is $10^{-4}$ for $W$ in BP, FA, DFA and SF. It is initially $0.03$ for $W$ and $R$ in PFA, but reduced to $10^{-4}$ after the first epoch. The networks are trained for 40 epochs with a batch size of 64 and 5 seeds. Training each network takes $\sim 10$ minutes on a GeForce RTX 3090.
 
\subsubsection{CIFAR10}
The learning rate is 0.1 for BP, FA, SF and PFA, and $10^{-4}$ for DFA, multiplied by a factor of 0.1 at the 100th, 150th and 200th epochs. The weight decay is $10^{-4}$ for $W$ in BP, FA, DFA and SF, and $0.005$ for $W$ and $R$ in PFA (reduced to $10^{-4}$ after the first epoch). We trained the networks for 200 epochs with a batch size of 128 and 5 seeds. Training each network takes $\sim 2$ hours on a GeForce RTX 3090. 

\subsubsection{ImageNet}
The learning rate is 0.1 for BP, SF and PFA, multiplied by a factor of 0.1 at the 20th, 40th and 60th epochs. The weight decay is $10^{-4}$ for $W$ in BP and SF, and $0.0005$ for $W$ and $R$ in PFA (reduced to $10^{-4}$ after the first epoch). We trained the networks for 75 epochs with a batch size of 256. Training each network takes $\sim 1$ week on a GeForce RTX 3090. 

\subsection{Weight correlation/alignment in the brain}
\label{sec:brain_corr}
We ran simulations based on experimental observations \cite{song2005highly}. We first sampled feedforward weights from a normal distribution. For 69\% of feedforward weights (unidirectionally connected neurons), we set the corresponding feedback weights to 0. For the remaining 31\% (bidirectionally connected neurons), we randomly sampled feedback weights that correlate with feedforward weights (correlation coefficient $R=0.36$). The weight correlation for all pairs of connected neurons is then $R\approx 0.2$, corresponding to a backward-forward weight alignment angle of $78^\circ$. This result is an approximate estimate and might change under different assumptions.

\subsection{Proofs for three propositions}
\label{sec:proof}

\paragraph{Proposition 1.} In PFA-o ($\lambda\le 1$) and PFA (in the limit $\lambda\rightarrow 0$), the path alignment angle $\angle(\bm{Vec}(W^T), \bm{Vec}(RB))$ converges to 0 after learning.

Proof: In PFA-o, $B$ is a semi-orthogonal matrix, so $B^T B=I$. For PFA ($\lambda\rightarrow 0$), as a direct corollary of Marchenko–Pastur law, we have $\Pr(B^T B=I)=1$, i.e., almost surely. Therefore, the approximately equal sign in Eq. \ref{eq:alignment} becomes an equal sign, and we have $\Delta (RB)=\Delta W^T$. For non-zero weight decay, the effect of the weight initialization diminishes after a sufficient number of training steps. The final weights are then the weighted average of past weight updates, thus we have $(RB)= W^T$.

\paragraph{Proposition 2.} In PFA-o ($\lambda\le 1$) and PFA (in the limit $\lambda\rightarrow 0$), the error $e_l^{PFA}$ at layer $l$ is fully aligned to $e_l^{BP}$ after the path alignment angle $\angle(\bm{Vec}(W^T), \bm{Vec}(RB))$ converges to 0, and thus the loss function is monotonically decreasing for infinitesimal learning rates.

Proof: From \textbf{Proposition 1}, we have $RB=W^T$ after the path alignment angle converges to 0. The error signals at the last layer $L$ are the same by definition: $e_L^{\rm PFA}=e_L^{\rm BP}$. By induction, if $e_{l+1}^{\rm PFA}=e_{l+1}^{\rm BP}$, then $e_{l}^{\rm PFA}=e_{l}^{\rm BP}$, following their definitions. Thus, the weight update $\Delta W$ specified by PFA is the same as the weight update $\Delta W$ specified by BP, leading to a monotonically decreasing loss for infinitesimal learning rates.

\paragraph{Proposition 3.} In PFA (with $\lambda>0$), the path alignment angle $\angle(\bm{Vec}(W^T), \bm{Vec}(RB))$ converges to $\arccos(1/\sqrt{(1+\lambda)})$ almost surely after learning.

Proof: From Eq. \ref{eq:alignment}, after the effect of the weight initialization has vanished due to weight decay, we have 
\begin{equation*}
    (RB)^T=B^T BW
\end{equation*}
where W is an $N_{l+1} \times N_l$ matrix,  B is an $\bar{N}_l \times N_{l+1}$ matrix, and $B^TB$ is not assumed to equal the identity.  The normalized inner product is
\begin{equation*}
(\bm{Vec}(W^T), \bm{Vec}(RB))=\Tr(W^T(RB)^T)/\sqrt{\Tr(W^TW)\Tr(B^TR^TRB)}.
\end{equation*}
The numerator $\Tr(W^T(RB)^T)=\Tr(W^TB^TBW)$ is a quadratic form of $W$. Because $B^TB$ (an $N_{l+1} \times N_{l+1}$ matrix) is positive definite, we consider its eigendecomposition $B^TB=P^T V P$, where $P$ is an orthogonal matrix, and $V$ is diagonal. If we assume that $W$ is sampled from the standard multivariate normal distribution with zero mean and identity covariance matrix (elements i.i.d. from the unit normal distribution), $U=PW$ still has a standard multivariate normal distribution. Then 
\begin{equation*}
\Tr(W^TB^TBW)=\Tr(U^TVU)=\sum_{j=1}^{N_{l+1}} v_j \chi^2_j,
\end{equation*}
where $v_j$ is the $j$-th diagonal element of $V$, and $\chi^2_j$ is a random variable sampled from the Chi-squared distribution with $N_l$ degrees of freedom. Each $\chi^2_j$ has a mean of $N_l$, and a standard deviation of $\sqrt{2N_l}$. When $N_l$ is large enough, $\chi^2_j$ is tightly distributed around and thus well approximated by its mean value. The denominator of the normalized inner product similarly involves a weighted sum of Chi-squared variables which can be replaced by their mean values when $N_l$ is large enough. Thus the normalized inner product $(\bm{Vec}(W^T), \bm{Vec}(RB))$ almost surely converges to ${\mathbb E} v / \sqrt{{\mathbb E} v^2}=1/\sqrt{1+\lambda}$, where the expectation is integrated over the eigenvalue density $\mu(v)$ of $B^TB$ (Eq. \ref{eq:MP}), leading to a path alignment angle $\angle(\bm{Vec}(W^T), \bm{Vec}(RB))=\arccos(1/\sqrt{(1+\lambda)})$.

\end{document}